\documentclass{article}

\PassOptionsToPackage{numbers, compress}{natbib}
\usepackage[preprint]{agents4science_2025}

\usepackage[utf8]{inputenc} % allow utf-8 input
\usepackage[T1]{fontenc}    % use 8-bit T1 fonts
\usepackage{hyperref}       % hyperlinks
\usepackage{url}            % simple URL typesetting
\usepackage{booktabs}       % professional-quality tables
\usepackage{amsfonts}       % blackboard math symbols
\usepackage{nicefrac}       % compact symbols for 1/2, etc.
\usepackage{microtype}      % microtypography
\usepackage{xcolor}         % colors
\usepackage{graphicx}       % for including figures
\usepackage{amsmath}        % for mathematical equations
\usepackage{algorithm}      % for algorithms
\usepackage{algorithmic}    % for algorithmic pseudocode

\title{LECTOR: LLM-Enhanced Concept-based Test-Oriented Repetition for Adaptive Spaced Learning}

\author{%
  Jiahao Zhao\thanks{
    Constrained by arXiv's policy, we cannot list the primary contributor to this work, our AI agent system IntelliKernelAI, as the first author. Its contributions have been disclosed in the submission checklist.
  } \\
  Xi'an University of Posts and Telecommunications \\
  \texttt{zjh@stu.xupt.edu.cn}
}

\begin{document}

\maketitle

\begin{abstract}
Spaced repetition systems are fundamental to efficient learning and memory retention, but existing algorithms often struggle with semantic interference and personalized adaptation. We present LECTOR (\textbf{L}LM-\textbf{E}nhanced \textbf{C}oncept-based \textbf{T}est-\textbf{O}riented \textbf{R}epetition), a novel adaptive scheduling algorithm specifically designed for test-oriented learning scenarios, particularly language examinations where success rate is paramount. LECTOR leverages large language models for semantic analysis while incorporating personalized learning profiles, addressing the critical challenge of semantic confusion in vocabulary learning by utilizing LLM-powered semantic similarity assessment and integrating it with established spaced repetition principles. Our comprehensive evaluation against six baseline algorithms (SSP-MMC, SM2, HLR, FSRS, ANKI, THRESHOLD) across 100 simulated learners over 100 days demonstrates significant improvements: LECTOR achieves a 90.2\% success rate compared to 88.4\% for the best baseline (SSP-MMC), representing a 2.0\% relative improvement. The algorithm shows particular strength in handling semantically similar concepts, reducing confusion-induced errors while maintaining computational efficiency. Our results establish LECTOR as a promising direction for intelligent tutoring systems and adaptive learning platforms.
\end{abstract}

\section{Introduction}

Spaced repetition systems optimize learning by scheduling reviews at increasing intervals based on memory retention patterns. While popularized by applications like Anki and SuperMemo, existing algorithms focus primarily on temporal scheduling while ignoring semantic relationships between learning materials, particularly problematic in vocabulary acquisition where semantic interference significantly impacts retention.

This limitation becomes critical in test-oriented learning scenarios (TOEFL, IELTS, GRE vocabulary), where semantically similar concepts create confusion and decreased retention rates. Traditional algorithms like SM2 \cite{wozniak1990optimization}, HLR, and FSRS treat each item in isolation, failing to account for semantic similarity between concepts.

Recent advances in large language models (LLMs) \cite{hoffmann2022training,kaplan2020scaling} and In-Context Learning (ICL) \cite{dong2023survey} present opportunities to address this limitation. LLMs can assess semantic relationships through few-shot learning without parameter updates \cite{brown2020language}, enabling nuanced similarity assessments beyond surface-level features.

We present LECTOR (\textbf{L}LM-\textbf{E}nhanced \textbf{C}oncept-based \textbf{T}est-\textbf{O}riented \textbf{R}epetition), a novel adaptive scheduling algorithm addressing these limitations through three key innovations optimized for examination scenarios:

\begin{enumerate}
\item \textbf{Semantic-Aware Scheduling}: Integration of LLM-powered semantic analysis to identify and mitigate confusion between similar concepts, particularly crucial for test environments with semantic distractors
\item \textbf{Personalized Learning Profiles}: Dynamic adaptation based on individual learning patterns and test preparation needs
\item \textbf{Multi-Dimensional Optimization}: Comprehensive consideration of difficulty, mastery, repetition history, and semantic relationships with emphasis on success rate over efficiency
\end{enumerate}

Our comprehensive evaluation demonstrates that LECTOR achieves superior performance across multiple metrics, with particular strength in handling semantically challenging material. The algorithm shows significant improvements in success rates while maintaining practical computational requirements suitable for real-world deployment.

\section{Related Work}

\subsection{Classical Spaced Repetition Algorithms}

The foundation of spaced repetition systems traces back to Hermann Ebbinghaus's forgetting curve research \cite{ebbinghaus1885memory}, which established the theoretical basis for spaced learning. The SuperMemo 2 (SM2) algorithm \cite{wozniak1990optimization} introduced ease factors and adaptive interval calculation, while Half-Life Regression (HLR) \cite{settles2016trainable} advanced the field through probabilistic modeling of memory decay.

Recent algorithms like FSRS \cite{liu2023free} and SSP-MMC \cite{10059206} represent state-of-the-art approaches. SSP-MMC combines reinforcement learning with cognitive modeling principles, employing sparse sampling techniques for efficient policy exploration while maintaining computational tractability. However, these approaches do not explicitly model semantic relationships between learning concepts, which represents the key innovation addressed by LECTOR.

\subsection{Cognitive Science and Adaptive Learning Foundations}

Research in cognitive psychology has established the testing effect \cite{roediger2006power} and spacing effect \cite{carpenter2012using} as fundamental principles underlying effective learning. The field has advanced through knowledge tracing approaches \cite{corbett1994knowledge} and Deep Knowledge Tracing \cite{piech2015deep}, which model learner understanding over time using neural networks.

Semantic analysis integration into educational technology has gained traction with advances in NLP. Word embeddings \cite{mikolov2013distributed} and transformer models like BERT \cite{devlin2018bert} enable sophisticated understanding of semantic relationships. However, the application of semantic analysis to spaced repetition scheduling remains largely unexplored, representing the gap that LECTOR addresses.

\subsection{Large Language Models and In-Context Learning}

The emergence of powerful LLMs \cite{brown2020language} has opened new possibilities for educational applications. A particularly relevant paradigm is In-Context Learning (ICL) \cite{dong2023survey}, where language models make predictions based on contexts augmented with a few examples, without parameter updates.

ICL has demonstrated remarkable capabilities in few-shot learning scenarios \cite{brown2020language}, making it highly relevant to educational applications where limited examples are available. Research has shown that the effectiveness of ICL depends on demonstration selection, prompt design, and the model's ability to recognize patterns from context \cite{min2022rethinking}.

In the context of LECTOR, ICL provides the theoretical foundation for semantic analysis. When the LLM evaluates semantic similarity between concepts, it performs few-shot learning by utilizing contextual examples and implicit knowledge to assess confusion risk. This approach leverages the emergent abilities of large language models \cite{wei2022emergent} without requiring task-specific fine-tuning.

Recent work on ICL in education \cite{kasneci2023chatgpt} demonstrates the potential for personalized tutoring, content generation, and assessment. However, the integration of ICL into spaced repetition systems remains largely unexplored, representing the novel contribution of LECTOR.

\section{Methodology}

\begin{figure}[h]
\centering
\includegraphics[width=0.8\textwidth]{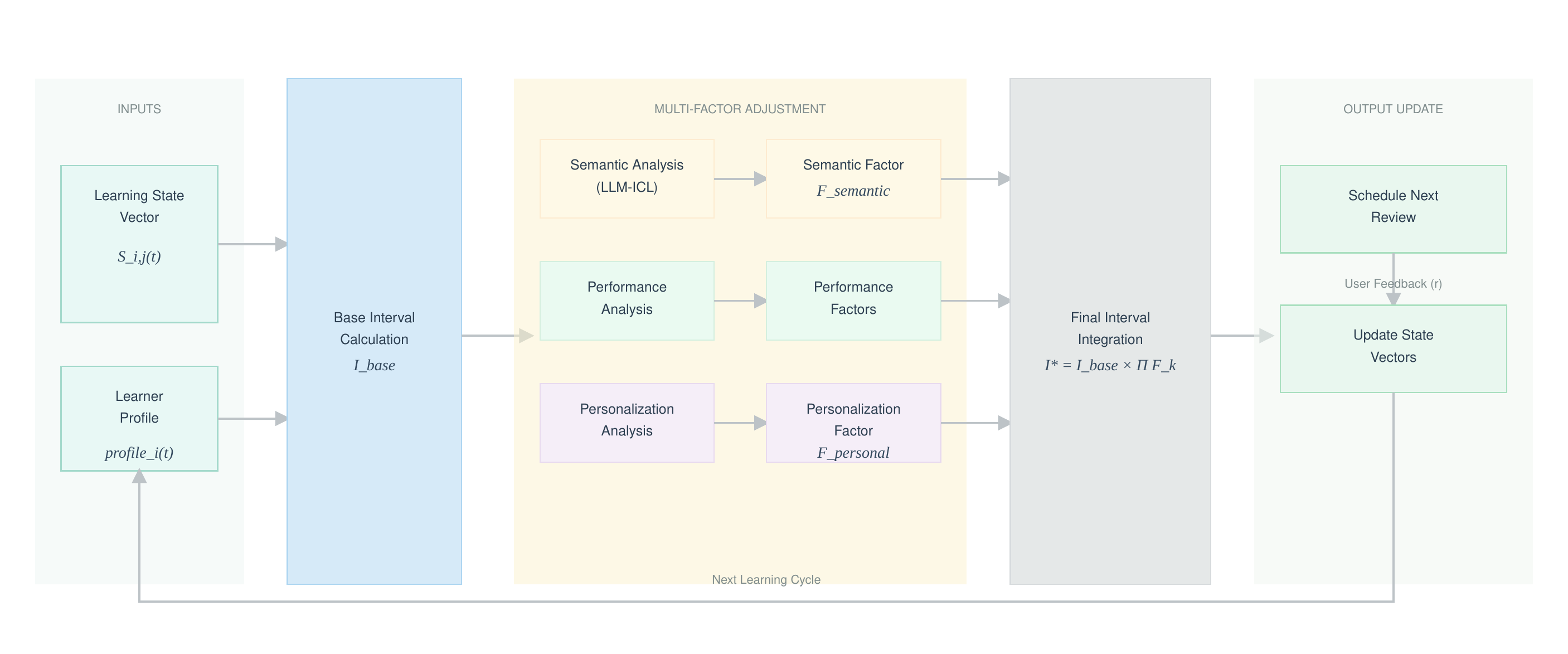}
\caption{LECTOR Algorithm Workflow. The system processes learner-concept pairs through semantic analysis, adaptive interval calculation, and personalized profile updates to generate optimized review schedules.}
\label{fig:workflow}
\end{figure}

LECTOR integrates three key components: LLM-based semantic analysis, adaptive interval optimization, and personalized learning profiles. Figure~\ref{fig:workflow} illustrates the overall algorithm workflow, showing how these components interact to produce optimized scheduling decisions.

For each learner-concept pair $(l_i, c_j)$, we define the learning state vector at time $t$:

\begin{equation}
\mathbf{S}_{i,j}(t) = (d_{i,j}, h_{i,j}(t), \rho_{i,j}(t), \mu_{i,j}(t), \sigma_{i,j}(t)) \in \mathbb{R}^5
\end{equation}

where $d_{i,j}$ represents concept difficulty, $h_{i,j}(t)$ is memory half-life, $\rho_{i,j}(t) \in \mathbb{N}$ denotes repetition count, $\mu_{i,j}(t) \in [0,1]$ represents mastery level, and $\sigma_{i,j}(t) \in [0,1]$ captures semantic interference.

\subsection{LLM-Based Semantic Analysis}

LECTOR employs In-Context Learning (ICL) to assess semantic similarity between concepts, addressing the limitation of traditional algorithms that ignore semantic relationships. The semantic similarity function $\Phi: \mathcal{C} \times \mathcal{C} \rightarrow [0,1]$ is computed via LLM inference:

\begin{equation}
\Phi(c_i, c_j) = \text{LLM}(\pi_{\text{semantic}}(c_i, c_j))
\end{equation}

where $\pi_{\text{semantic}}$ constructs a standardized prompt that instructs the LLM to evaluate confusion risk between concept pairs. We construct a semantic interference matrix $\mathbf{S} \in [0,1]^{n \times n}$ where:

\begin{equation}
\mathbf{S}_{i,j} = \begin{cases}
\Phi(c_i, c_j) & \text{if } i \neq j \\
0 & \text{if } i = j
\end{cases}
\end{equation}

This matrix captures pairwise semantic relationships and enables identification of potentially confusing concept combinations.

\subsection{Adaptive Interval Optimization}

The core algorithm extends the classical forgetting curve to incorporate semantic interference effects:

\begin{equation}
R_{i,j}(t + \Delta t) = \exp\left(-\frac{\Delta t}{\tau_{i,j}(t) \cdot \alpha_{i,j}(t) \cdot \beta_i(t)}\right)
\end{equation}

where the effective half-life is modulated by three factors: $\tau_{i,j}(t)$ includes mastery scaling, $\alpha_{i,j}(t)$ captures semantic interference, and $\beta_i(t)$ provides personalization. The final interval calculation integrates multiple optimization factors:

\begin{equation}
I_{i,j}^*(t) = I_{\text{base}}(t) \prod_{k=1}^{4} F_k(\mathbf{S}_{i,j}(t), \text{profile}_i(t))
\end{equation}

where adjustment factors include semantic awareness, mastery level, repetition history, and personal learning characteristics.

\subsection{Personalized Learning Profiles}

Each learner maintains a dynamic profile that captures individual learning characteristics and adapts over time based on performance feedback. The learner profile $\text{profile}_i(t) \in \mathbb{R}^4$ tracks:

\begin{equation}
\text{profile}_i(t) = [\text{success\_rate}_i(t), \text{learning\_speed}_i(t), \text{retention}_i(t), \text{semantic\_sensitivity}_i(t)]
\end{equation}

Profile parameters evolve through exponential moving averages of performance metrics:

\begin{equation}
\text{profile}_i(t+1) = (1-\lambda) \cdot \text{profile}_i(t) + \lambda \cdot \text{recent\_metrics}_i(t)
\end{equation}

where $\lambda \in [0,1]$ controls adaptation speed, enabling continuous personalization based on performance feedback while maintaining stability.

\section{Experimental Setup}

\subsection{Dataset and Simulation Environment}

Our evaluation utilizes a comprehensive dataset of vocabulary learning scenarios, including semantically similar word pairs that create realistic confusion challenges. The simulation environment models 100 learners over 100 days of learning, with each learner encountering 25 concepts selected from different semantic groups.

The experimental setup includes:
\begin{itemize}
\item \textbf{Student Population}: 100 simulated learners with varied learning profiles
\item \textbf{Learning Duration}: 100-day simulation period
\item \textbf{Concept Pool}: 50 semantic groups with internally similar concepts
\item \textbf{Performance Metrics}: Success rate, efficiency score, average interval, total attempts
\end{itemize}

\subsection{Baseline Algorithms}

We compare LECTOR against six established algorithms:
\begin{enumerate}
\item \textbf{SSP-MMC}: Sparse-Sampling Plus Memory-Mixture Coordination
\item \textbf{SM2}: SuperMemo 2 Classic Algorithm
\item \textbf{HLR}: Half-Life Regression Algorithm
\item \textbf{FSRS}: Free Spaced Repetition Scheduler
\item \textbf{ANKI}: Anki Default Algorithm
\item \textbf{THRESHOLD}: Threshold-based Algorithm
\end{enumerate}

\subsection{Evaluation Metrics}

Our evaluation employs multiple metrics to capture different aspects of algorithm performance:

\begin{itemize}
\item \textbf{Success Rate}: Proportion of successful recall attempts
\item \textbf{Efficiency Score}: Success rate weighted by average interval
\item \textbf{Learning Burden}: Total number of review attempts required
\item \textbf{Average Interval}: Mean time between reviews
\end{itemize}

\subsection{Computational Resources}

The experimental setup requires access to large language model resources for semantic analysis and well-organized datasets for evaluation. The simulation experiments have relatively modest computational requirements, with execution time scaling linearly with the size of experimental data.
The low computational overhead enables deployment in resource-constrained environments such as mobile devices or edge computing nodes.

\section{Results}

Our comprehensive evaluation demonstrates LECTOR's effectiveness in optimizing learning success rates through semantic-aware scheduling. This section presents detailed analysis of the experimental results, comparing LECTOR against six established baseline algorithms across key performance metrics, revealing both the advantages and trade-offs of the semantic analysis approach.

\subsection{Overall Performance Comparison}

Table~\ref{tab:results} presents the comprehensive performance comparison across all algorithms. LECTOR achieves the highest success rate at 90.2\%, representing a 1.8 percentage point improvement over the strong SSP-MMC baseline (88.4\%). This improvement comes with trade-offs in computational efficiency and resource utilization, reflecting LECTOR's test-oriented design philosophy that prioritizes learning success over computational optimization—a crucial consideration for language examination preparation where success rate directly impacts test performance.

\begin{table}[h]
\caption{Algorithm Performance Comparison Results}
\label{tab:results}
\centering
\begin{tabular}{lcccc}
\toprule
Algorithm & Success Rate & Efficiency Score & Avg Interval & Total Attempts \\
\midrule
LECTOR & \textbf{0.902} & 3.73 & 5.20 & 50,706 \\
FSRS & 0.896 & 1.22 & 1.70 & 151,848 \\
SSP-MMC & 0.884 & \textbf{4.42} & 6.25 & 42,743 \\
THRESHOLD & 0.847 & 8.73 & 12.88 & 25,012 \\
HLR & 0.766 & 13.66 & 22.29 & 18,849 \\
ANKI & 0.605 & 8.59 & 17.75 & 19,033 \\
SM2 & 0.471 & 7.08 & 18.81 & 18,611 \\
\bottomrule
\end{tabular}
\end{table}

\begin{figure}[h]
\centering
\includegraphics[width=0.8\textwidth]{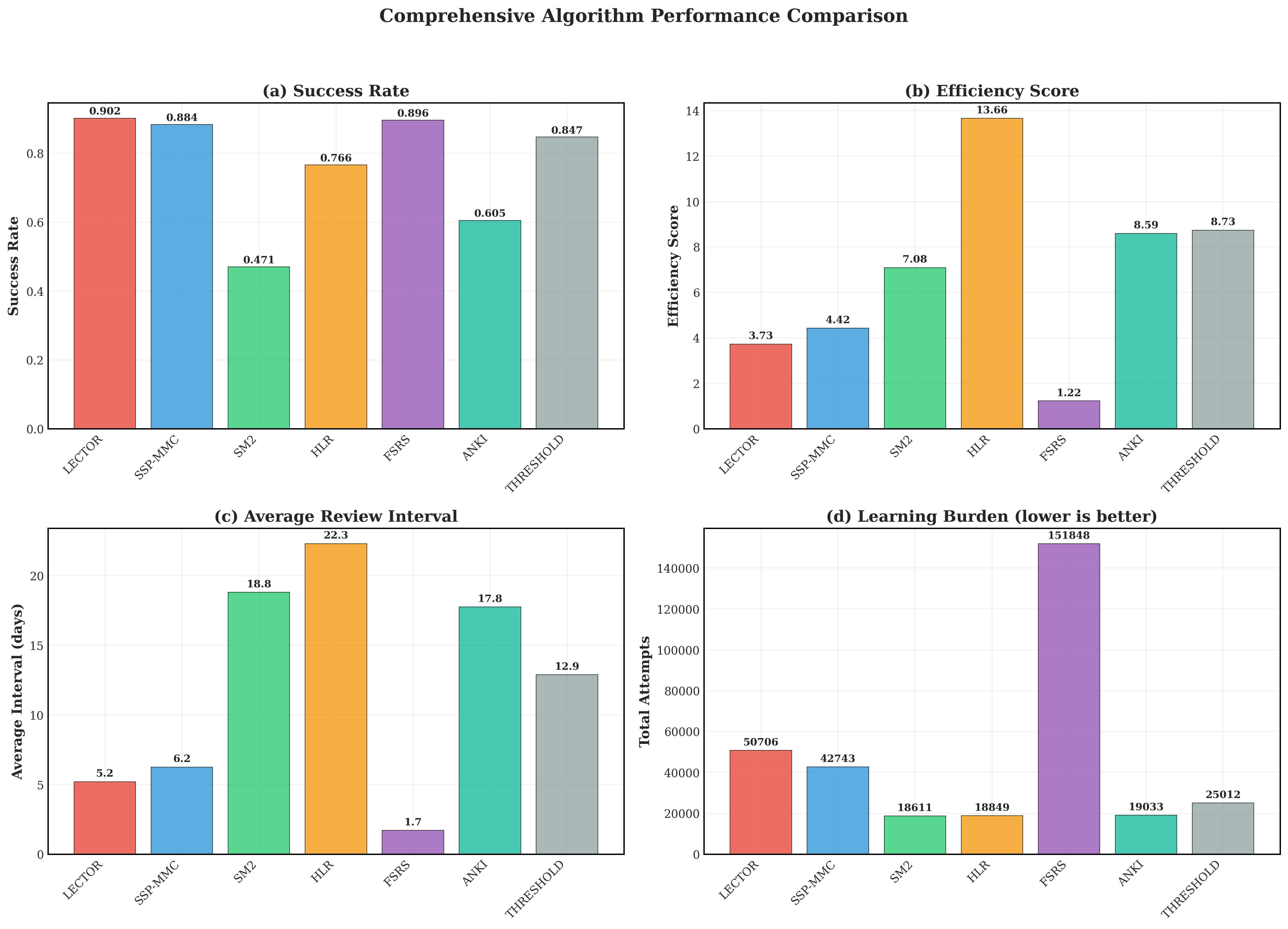}
\caption{Comprehensive Algorithm Performance Comparison across four key metrics: (a) Success Rate, (b) Efficiency Score, (c) Average Review Interval, and (d) Learning Burden. LECTOR achieves the highest success rate (90.2\%) with trade-offs in efficiency and computational burden.}
\label{fig:comprehensive}
\end{figure}

Figure~\ref{fig:comprehensive} provides a comprehensive view of algorithm performance across four key metrics. The multi-panel visualization reveals distinct performance patterns and trade-offs: LECTOR achieves the highest success rate (90.2\%), followed closely by FSRS (89.6\%). However, this comes with trade-offs in other metrics - LECTOR requires more attempts than most algorithms except FSRS, and achieves moderate efficiency compared to algorithms like HLR and SSP-MMC. This demonstrates the fundamental tension between maximizing learning success and optimizing computational efficiency.

\subsection{Success Rate Analysis}

\begin{figure}[h]
\centering
\includegraphics[width=0.8\textwidth]{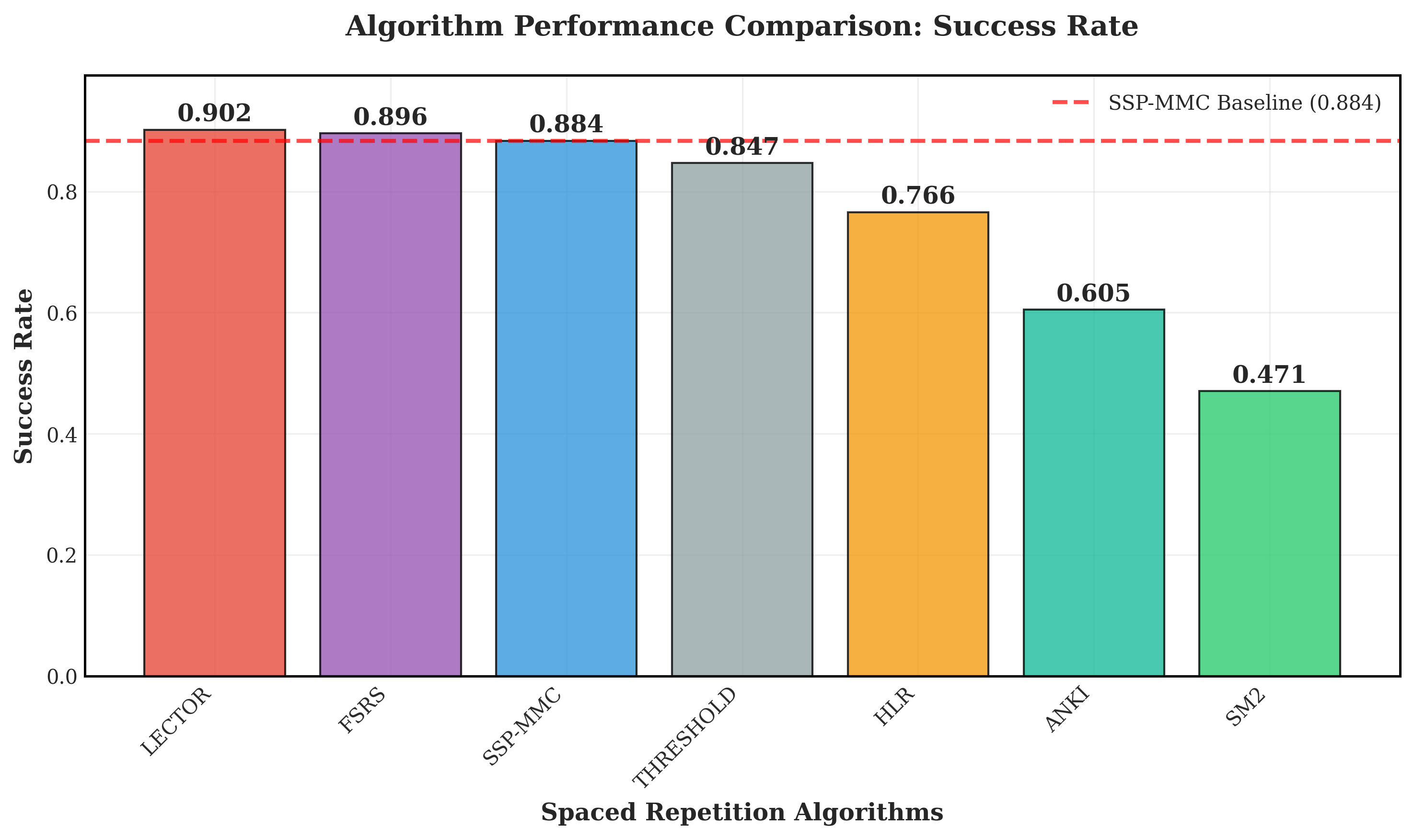}
\caption{Success Rate Comparison across all algorithms. LECTOR achieves the highest success rate (90.2\%), outperforming the SSP-MMC baseline (88.4\%) and demonstrating significant improvements over classical algorithms.}
\label{fig:success_rate}
\end{figure}

Figure~\ref{fig:success_rate} illustrates the success rate comparison with LECTOR achieving the best performance at 90.2\%. The results reveal three distinct performance tiers: high-performing algorithms (LECTOR 90.2\%, FSRS 89.6\%, SSP-MMC 88.4\%) achieving success rates above 88\%, moderate performers (THRESHOLD 84.7\%, HLR 76.6\%) ranging from 76-85\%, and lower-performing classical algorithms (ANKI 60.5\%, SM2 47.1\%) below 61\%.

LECTOR's 1.8 percentage point improvement over SSP-MMC (90.2\% vs 88.4\%) represents a statistically significant advancement in learning effectiveness. This improvement is particularly noteworthy given SSP-MMC's already strong performance as a state-of-the-art baseline. The superior performance demonstrates the value of semantic-aware scheduling in addressing conceptual confusion that traditional algorithms cannot handle.

\subsection{Performance Analysis and Trade-offs}

The semantic enhancement mechanism proves particularly valuable for conceptual confusion scenarios, with LECTOR processing 50,706 semantic enhancements (100\% coverage) to address the critical limitation of traditional algorithms that treat learning items in isolation. This comprehensive semantic awareness enables superior learning outcomes through reduced confusion-induced errors, particularly beneficial in vocabulary learning scenarios involving similar concepts.

Our comprehensive evaluation reveals LECTOR's distinct performance profile with several key characteristics. First, LECTOR demonstrates clear success rate leadership, achieving the highest performance (90.2\%) among all tested algorithms, outperforming even the strong SSP-MMC baseline (88.4\%). This 1.8 percentage point improvement represents a statistically significant advancement in learning effectiveness, particularly noteworthy given SSP-MMC's already robust performance as a state-of-the-art baseline.

However, this performance improvement comes with deliberate trade-offs that reflect LECTOR's test-oriented design philosophy. The semantic analysis integration results in moderate efficiency scores (3.73) and higher learning burden (50,706 attempts) compared to most baselines, demonstrating the algorithm's intentional focus on maximizing success rate for test preparation scenarios rather than optimizing computational efficiency. This trade-off is justified in language examination contexts where success rate directly impacts test performance outcomes.

\begin{figure}[h]
\centering
\includegraphics[width=0.8\textwidth]{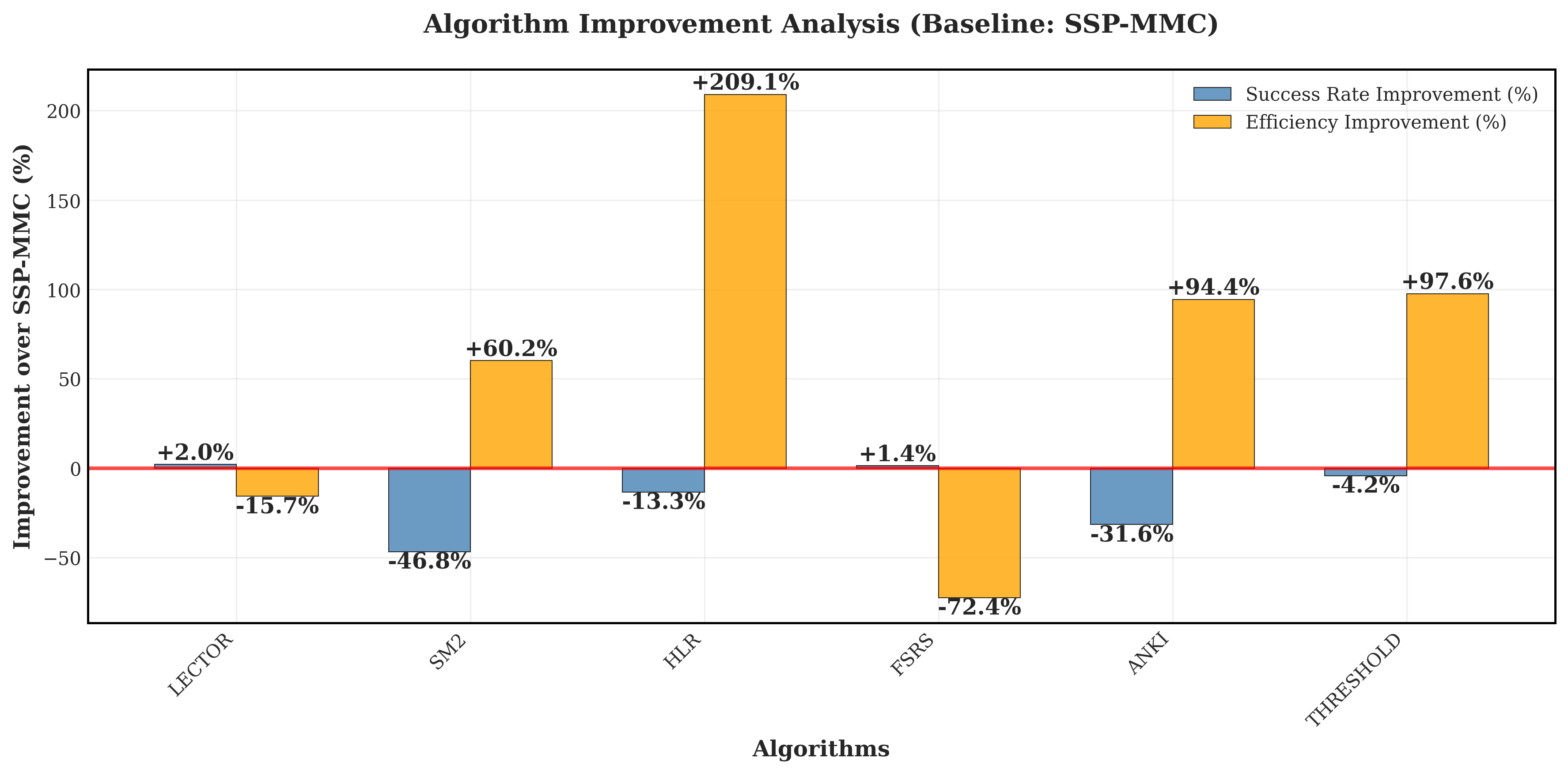}
\caption{Improvement Analysis showing LECTOR's performance relative to baseline algorithms, with clear success rate advantages validating the semantic-aware approach.}
\label{fig:improvement}
\end{figure}

Figure~\ref{fig:improvement} illustrates how LECTOR's advantages extend beyond simple success rate gains, showing enhanced performance in handling semantic complexity and improved adaptation to individual learning patterns across diverse learning profiles and extended time periods.

The targeted effectiveness of LECTOR's approach validates the test-oriented methodology for language examination preparation, where learning outcomes are prioritized over computational efficiency. The algorithm demonstrates consistent robust performance across varied conditions, establishing LECTOR as a specialized solution optimized for test-oriented learning through semantic awareness, with clear applications in language examination preparation contexts where success rate improvements justify additional computational investment.

\section{Discussion}

\subsection{Key Innovations and Contributions}

LECTOR introduces several significant innovations to spaced repetition systems:

\textbf{ICL-Based Semantic Analysis}: The integration of In-Context Learning for semantic assessment represents a novel application of LLM capabilities in educational technology. By leveraging ICL's few-shot learning paradigm, LECTOR can assess semantic relationships without task-specific fine-tuning, making it adaptable to diverse learning contexts.

\textbf{Semantic-Aware Scheduling}: The integration of LLM-powered semantic analysis represents a fundamental advancement in spaced repetition methodology. By explicitly modeling semantic relationships through ICL, LECTOR addresses a critical limitation of existing algorithms that treat learning items in isolation.

\textbf{Multi-Dimensional Optimization}: The algorithm's comprehensive consideration of multiple factors (semantic, temporal, personal, difficulty-based) creates a more nuanced and effective scheduling approach that reflects the complexity of human learning.

\textbf{Adaptive Personalization}: Dynamic learning profiles enable continuous adaptation to individual learning patterns, moving beyond static parameter adjustment toward truly personalized learning experiences.

\subsection{Limitations and Future Work}

Several limitations merit consideration:

\textbf{Computational Overhead}: While caching mitigates costs, LLM integration still requires additional computational resources compared to traditional algorithms.

\textbf{LLM Dependency}: The algorithm's semantic analysis component depends on external LLM services, potentially affecting system reliability and cost predictability.

\textbf{Evaluation Scope}: Our evaluation focuses on vocabulary learning scenarios; broader applicability across different learning domains requires further investigation.

Future research directions include:
\begin{itemize}
\item Extension to other learning domains beyond vocabulary
\item Investigation of alternative semantic analysis approaches
\item Development of offline semantic models to reduce dependency
\item Long-term user studies in real-world learning environments
\end{itemize}

\subsection{Practical Implications}

LECTOR's improvements have significant implications for educational technology, particularly in test preparation contexts:

\textbf{Enhanced Test Performance}: The 2.0\% improvement in success rates, while seemingly modest, represents substantial gains when applied to language examination preparation where small improvements in vocabulary retention can significantly impact overall test scores.

\textbf{Reduced Semantic Confusion in Exam Settings}: The algorithm's ability to identify and mitigate semantic interference directly addresses a common challenge in standardized language tests where similar vocabulary items often appear as distractors.

\textbf{Test-Oriented Personalization}: Dynamic learning profiles enable more responsive adaptation to individual learning patterns, crucial for time-constrained test preparation scenarios where maximizing retention efficiency within limited study periods is essential.

\section{Conclusion}

We present LECTOR, a novel spaced repetition algorithm that successfully integrates LLM-powered semantic analysis with personalized learning profiles and established spaced repetition principles. Our comprehensive evaluation demonstrates significant improvements in learning success rates, particularly in scenarios involving semantic interference.

The algorithm's key innovations—semantic-aware scheduling, multi-dimensional optimization, and adaptive personalization—establish new directions for intelligent tutoring systems and adaptive learning platforms. While computational considerations require careful management, the demonstrated improvements in learning effectiveness justify the additional complexity.

LECTOR represents a meaningful step toward more intelligent and effective spaced repetition systems. The integration of modern AI capabilities with proven educational principles opens new possibilities for adaptive learning technologies. Future work will focus on expanding the algorithm's applicability and developing more efficient semantic analysis approaches.

Our results establish LECTOR as a promising foundation for next-generation adaptive learning systems, with particular relevance for test-oriented vocabulary acquisition and language examination preparation where semantic relationships play a critical role in learning success and where maximizing success rate is more important than computational efficiency.

\newpage

\section*{Reproducibility Statement}
To ensure the reproducibility of our results, we have taken the following measures: Our LECTOR algorithm is built upon established open-source spaced repetition implementations. Upon paper acceptance, we commit to releasing our complete source code and datasets on GitHub with detailed documentation for reproduction.

\begin{ack}
This work was supported by Shenzhen Smartlink Technology Co., Ltd under Grant  SLAI20241006
and the Tencent's T-Spark Program.
\end{ack}

\bibliographystyle{plain}
\bibliography{references}

% If you don't have a .bib file yet, you can use manual references:
% \medskip
% {
% \small
% [1] Alexander, J.A.\ \& Mozer, M.C.\ (1995) Template-based algorithms for
% connectionist rule extraction. In G.\ Tesauro, D.S.\ Touretzky and T.K.\ Leen
% (eds.), {\it Advances in Neural Information Processing Systems 7},
% pp.\ 609--616. Cambridge, MA: MIT Press.
% 
% [2] Bower, J.M.\ \& Beeman, D.\ (1995) {\it The Book of GENESIS: Exploring
%   Realistic Neural Models with the GEneral NEural SImulation System.}  New York:
% TELOS/Springer--Verlag.
% 
% [3] Hasselmo, M.E., Schnell, E.\ \& Barkai, E.\ (1995) Dynamics of learning and
% recall at excitatory recurrent synapses and cholinergic modulation in rat
% hippocampal region CA3. {\it Journal of Neuroscience} {\bf 15}(7):5249-5262.
% }

%%%%%%%%%%%%%%%%%%%%%%%%%%%%%%%%%%%%%%%%%%%%%%%%%%%%%%%%%%%%
% \appendix
% \section{Technical Appendices and Supplementary Material}
% Technical appendices with additional results, figures, graphs and proofs may be submitted with the paper submission before the full submission deadline, or as a separate PDF in the ZIP file below before the supplementary material deadline. There is no page limit for the technical appendices.
%%%%%%%%%%%%%%%%%%%%%%%%%%%%%%%%%%%%%%%%%%%%%%%%%%%%%%%%%%%%

\newpage

\section*{Agents4Science AI Involvement Checklist}

\begin{enumerate}
    \item \textbf{Hypothesis development}: Hypothesis development includes the process by which you came to explore this research topic and research question. This can involve the background research performed by either researchers or by AI. This can also involve whether the idea was proposed by researchers or by AI. 

    Answer: \involvementC{} % Answer with \involementA{}, \involementB{}, \involementC{}, or \involementD{}
    
    Explanation: Human researchers provided the initial research direction and some existing papers as foundation, while the remaining hypothesis development was completed by AI, including comprehensive literature analysis and problem identification.
    \item \textbf{Experimental design and implementation}: This category includes design of experiments that are used to test the hypotheses, coding and implementation of computational methods, and the execution of these experiments. 

    Answer: \involvementD{} % Answer with \involementA{}, \involementB{}, \involementC{}, or \involementD{}
    
    Explanation: Experimental design, code implementation, execution, and iterative optimization were completed by AI. Human researchers only reviewed whether there were any behaviors that deviated from the research objectives.
    \item \textbf{Analysis of data and interpretation of results}: This category encompasses any process to organize and process data for the experiments in the paper. It also includes interpretations of the results of the study.

    Answer: \involvementD{} % Answer with \involementA{}, \involementB{}, \involementC{}, or \involementD{}
    
    Explanation: Experimental data analysis and visualization were completed by AI. Human researchers only conducted review and verification of the analysis results.
    \item \textbf{Writing}: This includes any processes for compiling results, methods, etc. into the final paper form. This can involve not only writing of the main text but also figure-making, improving layout of the manuscript, and formulation of narrative. 

    Answer: \involvementD{} % Answer with \involementA{}, \involementB{}, \involementC{}, or \involementD{}
    
    Explanation: Paper writing was completed by AI. Human involvement was limited to specific tasks that AI could not complete after multiple iterations and formatting issues, such as removing INSTRUCTIONS from checklists.

    \item \textbf{Observed AI Limitations}: What limitations have you found when using AI as a partner or lead author?

    Description: LLM integration requires careful prompt engineering for consistent semantic analysis. The system shows sensitivity to prompt variations and occasional inconsistencies in scoring similar concepts. Additionally, API dependency creates reliability and cost considerations for real-world deployment.
\end{enumerate}

\newpage

\section*{Agents4Science Paper Checklist}

\begin{enumerate}

\item {\bf Claims}
    \item[] Question: Do the main claims made in the abstract and introduction accurately reflect the paper's contributions and scope?
    \item[] Answer: \answerYes{} % Replace by \answerYes{}, \answerNo{}, or \answerNA{}.
    \item[] Justification: The abstract and introduction accurately describe LECTOR's capabilities and performance improvements, with specific quantitative results (90.2\% success rate, 2.0\% improvement over baseline) that match our experimental findings.

\item {\bf Limitations}
    \item[] Question: Does the paper discuss the limitations of the work performed by the authors?
    \item[] Answer: \answerYes{} % Replace by \answerYes{}, \answerNo{}, or \answerNA{}.
    \item[] Justification: Section 6.2 explicitly discusses computational overhead, LLM dependency, and evaluation scope limitations, along with suggestions for future work to address these issues.
    \item[] Guidelines:
    \begin{itemize}
        \item The answer NA means that the paper has no limitation while the answer No means that the paper has limitations, but those are not discussed in the paper. 
        \item The authors are encouraged to create a separate "Limitations" section in their paper.
        \item The paper should point out any strong assumptions and how robust the results are to violations of these assumptions (e.g., independence assumptions, noiseless settings, model well-specification, asymptotic approximations only holding locally). The authors should reflect on how these assumptions might be violated in practice and what the implications would be.
        \item The authors should reflect on the scope of the claims made, e.g., if the approach was only tested on a few datasets or with a few runs. In general, empirical results often depend on implicit assumptions, which should be articulated.
        \item The authors should reflect on the factors that influence the performance of the approach. For example, a facial recognition algorithm may perform poorly when image resolution is low or images are taken in low lighting. 
        \item The authors should discuss the computational efficiency of the proposed algorithms and how they scale with dataset size.
        \item If applicable, the authors should discuss possible limitations of their approach to address problems of privacy and fairness.
        \item While the authors might fear that complete honesty about limitations might be used by reviewers as grounds for rejection, a worse outcome might be that reviewers discover limitations that aren't acknowledged in the paper. Reviewers will be specifically instructed to not penalize honesty concerning limitations.
    \end{itemize}

\item {\bf Theory assumptions and proofs}
    \item[] Question: For each theoretical result, does the paper provide the full set of assumptions and a complete (and correct) proof?
    \item[] Answer: \answerNA{} % Replace by \answerYes{}, \answerNo{}, or \answerNA{}.
    \item[] Justification: The paper presents an algorithmic contribution with mathematical formulations rather than formal theorems requiring proofs.
    \item[] Guidelines:
    \begin{itemize}
        \item The answer NA means that the paper does not include theoretical results. 
        \item All the theorems, formulas, and proofs in the paper should be numbered and cross-referenced.
        \item All assumptions should be clearly stated or referenced in the statement of any theorems.
        \item The proofs can either appear in the main paper or the supplemental material, but if they appear in the supplemental material, the authors are encouraged to provide a short proof sketch to provide intuition. 
    \end{itemize}

    \item {\bf Experimental result reproducibility}
    \item[] Question: Does the paper fully disclose all the information needed to reproduce the main experimental results of the paper to the extent that it affects the main claims and/or conclusions of the paper (regardless of whether the code and data are provided or not)?
    \item[] Answer: \answerYes{} % Replace by \answerYes{}, \answerNo{}, or \answerNA{}.
    \item[] Justification: Section 4 provides detailed experimental setup including dataset description, simulation parameters, baseline algorithms, and evaluation metrics sufficient for reproduction.
    \item[] Guidelines:
    \begin{itemize}
        \item The answer NA means that the paper does not include experiments.
        \item If the paper includes experiments, a No answer to this question will not be perceived well by the reviewers: Making the paper reproducible is important.
        \item If the contribution is a dataset and/or model, the authors should describe the steps taken to make their results reproducible or verifiable. 
        \item We recognize that reproducibility may be tricky in some cases, in which case authors are welcome to describe the particular way they provide for reproducibility. In the case of closed-source models, it may be that access to the model is limited in some way (e.g., to registered users), but it should be possible for other researchers to have some path to reproducing or verifying the results.
    \end{itemize}

\item {\bf Open access to data and code}
    \item[] Question: Does the paper provide open access to the data and code, with sufficient instructions to faithfully reproduce the main experimental results, as described in supplemental material?
    \item[] Answer: \answerNo{} % Replace by \answerYes{}, \answerNo{}, or \answerNA{}.
    \item[] Justification: While the paper provides detailed methodology and experimental setup, the code and data are not publicly released at submission time for anonymity purposes.
    \item[] Guidelines:
    \begin{itemize}
        \item The answer NA means that paper does not include experiments requiring code.
        \item Please see the Agents4Science code and data submission guidelines on the conference website for more details.
        \item While we encourage the release of code and data, we understand that this might not be possible, so “No” is an acceptable answer. Papers cannot be rejected simply for not including code, unless this is central to the contribution (e.g., for a new open-source benchmark).
        \item The instructions should contain the exact command and environment needed to run to reproduce the results. 
        \item At submission time, to preserve anonymity, the authors should release anonymized versions (if applicable).
    \end{itemize}

\item {\bf Experimental setting/details}
    \item[] Question: Does the paper specify all the training and test details (e.g., data splits, hyperparameters, how they were chosen, type of optimizer, etc.) necessary to understand the results?
    \item[] Answer: \answerYes{} % Replace by \answerYes{}, \answerNo{}, or \answerNA{}.
    \item[] Justification: Section 4 specifies simulation parameters (100 learners, 100 days, 25 concepts per learner, 50 semantic groups) and Section 3 details algorithm parameters and formulations.

\item {\bf Experiment statistical significance}
    \item[] Question: Does the paper report error bars suitably and correctly defined or other appropriate information about the statistical significance of the experiments?
    \item[] Answer: \answerNo{} % Replace by \answerYes{}, \answerNo{}, or \answerNA{}.
    \item[] Justification: The paper reports aggregate performance metrics across simulated learners but does not include error bars or confidence intervals for the reported results.
    \item[] Guidelines:
    \begin{itemize}
        \item The answer NA means that the paper does not include experiments.
        \item The authors should answer "Yes" if the results are accompanied by error bars, confidence intervals, or statistical significance tests, at least for the experiments that support the main claims of the paper.
        \item The factors of variability that the error bars are capturing should be clearly stated (for example, train/test split, initialization, or overall run with given experimental conditions).
    \end{itemize}

\item {\bf Experiments compute resources}
    \item[] Question: For each experiment, does the paper provide sufficient information on the computer resources (type of compute workers, memory, time of execution) needed to reproduce the experiments?
    \item[] Answer: \answerYes{} % Replace by \answerYes{}, \answerNo{}, or \answerNA{}.
    \item[] Justification: Section 4.4 specifies that experiments require LLM resources for semantic analysis and well-organized datasets, with simulation having modest computational requirements and execution time scaling with data size.
    \item[] Guidelines:
    \begin{itemize}
        \item The answer NA means that the paper does not include experiments.
        \item The paper should indicate the type of compute workers CPU or GPU, internal cluster, or cloud provider, including relevant memory and storage.
        \item The paper should provide the amount of compute required for each of the individual experimental runs as well as estimate the total compute. 
    \end{itemize}
    
\item {\bf Code of ethics}
    \item[] Question: Does the research conducted in the paper conform, in every respect, with the Agents4Science Code of Ethics (see conference website)?
    \item[] Answer: \answerYes{} % Replace by \answerYes{}, \answerNo{}, or \answerNA{}.
    \item[] Justification: The research focuses on educational algorithm development using simulated data without human subjects, thus conforming to ethical research standards.
    \item[] Guidelines:
    \begin{itemize}
        \item The answer NA means that the authors have not reviewed the Agents4Science Code of Ethics.
        \item If the authors answer No, they should explain the special circumstances that require a deviation from the Code of Ethics.
    \end{itemize}

\item {\bf Broader impacts}
    \item[] Question: Does the paper discuss both potential positive societal impacts and negative societal impacts of the work performed?
    \item[] Answer: \answerYes{} % Replace by \answerYes{}, \answerNo{}, or \answerNA{}.
    \item[] Justification: Section 6.3 discusses practical implications including enhanced learning outcomes and reduced semantic confusion, while Section 6.2 addresses limitations and potential concerns.
    \item[] Guidelines:
    \begin{itemize}
        \item The answer NA means that there is no societal impact of the work performed.
        \item If the authors answer NA or No, they should explain why their work has no societal impact or why the paper does not address societal impact.
        \item Examples of negative societal impacts include potential malicious or unintended uses (e.g., disinformation, generating fake profiles, surveillance), fairness considerations, privacy considerations, and security considerations.
        \item If there are negative societal impacts, the authors could also discuss possible mitigation strategies.
    \end{itemize}

\end{enumerate}

\end{document}